\pdfoutput=1

\documentclass[11pt]{article}

\usepackage[final]{acl}
\usepackage[english]{babel}

\usepackage{times}
\usepackage{latexsym}
\usepackage{arydshln}

\usepackage[T1]{fontenc}

\usepackage[utf8]{inputenc}

\usepackage{microtype}

\usepackage{inconsolata}

\usepackage{times}
\usepackage{latexsym}

\usepackage[T1]{fontenc}

\usepackage[utf8]{inputenc}

\usepackage{microtype}
\usepackage{graphicx}
\usepackage{subfigure}
\usepackage{inconsolata}
\usepackage{multirow}
\usepackage{CJKutf8}
\usepackage{enumitem}
\usepackage{amsmath}
\usepackage{float}

\title{A Novel Paradigm Boosting Translation Capabilities of Large Language Models}

\author{Jiaxin Guo\thanks{Co-First Author}, Hao Yang\thanks{Co-First Author}, Zongyao Li\\ 
        {\bf Daimeng Wei, Hengchao Shang, Xiaoyu Chen} \\
        \{jiaxinguo1,yanghao30,lizongyao\}@huawei.com\\
        \{weidaimeng,shanghengchao,chenxiaoyu35\}@huawei.com\\
        Huawei Translation Services Center, Beijing, China }

\begin{document}
\maketitle
\begin{abstract}
This paper presents a study on strategies to enhance the translation capabilities of large language models (LLMs) in the context of machine translation (MT) tasks. The paper proposes a novel paradigm consisting of three stages: Secondary Pre-training using Extensive Monolingual Data, Continual Pre-training with Interlinear Text Format Documents, and Leveraging Source-Language Consistent Instruction for Supervised Fine-Tuning. Previous research on LLMs focused on various strategies for supervised fine-tuning (SFT), but their effectiveness has been limited. While traditional machine translation approaches rely on vast amounts of parallel bilingual data, our paradigm highlights the importance of using smaller sets of high-quality bilingual data. We argue that the focus should be on augmenting LLMs' cross-lingual alignment abilities during pre-training rather than solely relying on extensive bilingual data during SFT. Experimental results conducted using the Llama2\cite{touvron2023llama}
model, particularly on Chinese-Llama2\cite{Chinese-LLaMA-Alpaca} after monolingual augmentation, demonstrate the improved translation capabilities of LLMs. A significant contribution of our approach lies in Stage2: Continual Pre-training with Interlinear Text Format Documents, which requires less than 1B training data, making our method highly efficient. Additionally, in Stage3, we observed that setting instructions consistent with the source language benefits the supervised fine-tuning process. Experimental results demonstrate that our approach surpasses previous work and achieves superior performance compared to models such as NLLB-54B\cite{nllbteam2022language} and GPT3.5-text-davinci-003, despite having a significantly smaller parameter count of only 7B or 13B. This achievement establishes our method as a pioneering strategy in the field of machine translation.
\end{abstract}

\section{Introduction}

Translation capabilities of large language models (LLMs)\cite{DBLP:journals/corr/abs-2005-14165,DBLP:journals/jmlr/ChowdheryNDBMRBCSGSSTMRBTSPRDHPBAI23,touvron2023llama} in machine translation (MT) tasks have been explored extensively in previous research\cite{jiao2023parrot,zeng2023tim,chen2023improving,xu2023paradigm,yang2023bigtranslate,zhang2023bayling}. However, achieving significant improvements in translation quality through supervised fine-tuning (SFT) strategies has proven challenging. Traditionally, machine translation relies on vast amounts of parallel bilingual data, but SFT only requires a small amount of high-quality bilingual data, highlighting a crucial distinction. It is a naive approach to consider using vast quantities of parallel bilingual data during SFT. However, experiments have shown that increasing the data volume yields limited improvements in quality and even leads to performance degradation on certain test sets. \textbf{Thus, the question arises: are extensive parallel bilingual data useless in SFT, or are they being misused?}

In this paper, we propose a novel training paradigm, consisting of three stages, to boost the translation capabilities of LLMs. Our contributions include refining the training strategy for downstream tasks and emphasizing the enhancement of LLMs' cross-lingual alignment abilities during pre-training. These contributions address the limitations observed in previous research. Our training paradigm comprises the following stages:

\textbf{Stage 1: Continual Pre-training using Extensive Monolingual Data.}
Consistent with previous findings\cite{xu2023paradigm}, we validate the effectiveness of monolingual data augmentation. Specifically, we perform SFT on a Chinese-Llama2\cite{Chinese-LLaMA-Alpaca} model, which undergoes monolingual data augmentation, thereby demonstrating improved translation capabilities.

\textbf{Stage 2: Continual Pre-training with Sentence-aligned Parallel Data.}
We construct interlinear text format from sentence-aligned bilingual parallel data and utilize them for continual pre-training of LLMs. Experimental results demonstrate the critical importance of this stage, resulting in a significant improvement in translation quality, particularly for English-Other translations. Stage 2 stands as a pivotal contribution in our paper, requiring less than 1B training data, thereby enhancing training efficiency.

\textbf{Stage 3: Leveraging Source-Language Consistent Instruction for Supervised Fine-Tuning.} In SFT, we discover that using instruction aligned with the source language of the translation notably improves performance. Leveraging source-language consistent instructions during SFT yields significant enhancements.

In summary, we introduce a three-stage training paradigm, highlighting the effectiveness of secondary pre-training, continual pre-training with interlinear text format documents, and leveraging source-language consistent instruction for supervised fine-tuning. These contributions address the limitations observed in previous research and pave the way for improved translation quality.

\section{Related Work}

\begin{figure*}
\centering
\includegraphics[width=15cm]{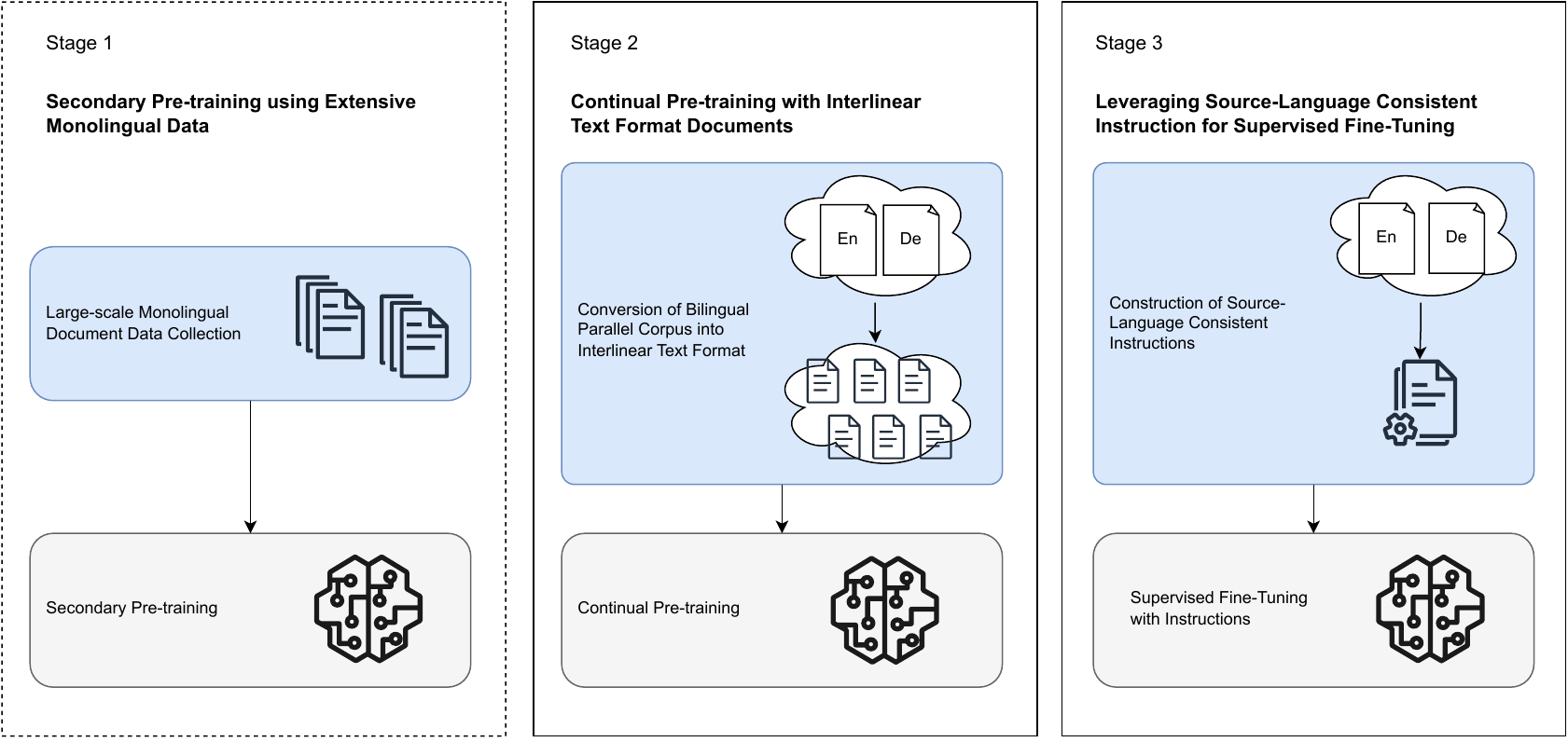}
\caption{The overall of our approach. Stage 1: Secondary Pre-training using Extensive Monolingual Data. Stage 2: Continual Pre-training with Interlinear Text Format Documents. Stage 3: Leveraging Source-Language Consistent Instruction for Supervised Fine-Tuning. *It should be noted that Stage 1 is considered non-essential.*}
\label{figure:model}
\end{figure*}

\subsection{Large Language Models}
\paragraph{Foundation Model} Foundation Model, a product of pre-training, is a prominent type of Large Language Model. It has gained substantial recognition in recent years for its impressive capabilities in natural language processing tasks. The most prevalent architectural framework for such models is the Transformer, which employs a series of self-attention mechanisms to process input text efficiently.

Among the state-of-the-art Large Language Models, notable examples include GPT-3\cite{DBLP:journals/corr/abs-2005-14165} and Llama2\cite{touvron2023llama}. These models have been widely lauded for their exceptional proficiency in understanding and generating natural language text. They showcase the remarkable potential of Foundation Models, pushing the boundaries of language processing and setting new benchmarks in various applications.

\paragraph{Instruct/Chat Model}
Instruct/Chat Model, a variant of Large Language Models, is specifically developed through the process of Supervised Fine-Tuning (SFT). Unlike Foundation Models, which are pre-trained, Instruct/Chat Models undergo additional supervised training to enhance their performance in specific tasks such as instruction following or conversational dialogue.

Supervised Fine-Tuning involves training the model on labeled datasets, where human annotators provide examples of desired input-output behavior. This approach enables Instruct/Chat Models to learn task-specific skills and exhibit improved performance in situations that require language understanding, generation, and interaction.

Noteworthy advancements have been observed in Instruct/Chat Models, with notable examples including models like ChatGPT. These models have exhibited remarkable outcomes in conversational scenarios, demonstrating their potential in enabling interactive and engaging human-like conversations.

\subsection{Machine Translation Task}

Machine Translation Task refers to the process of automatically translating text from one language to another using computational methods.



\paragraph{Traditional Methods} Traditional machine translation methods primarily rely on encoder-decoder\cite{DBLP:conf/nips/VaswaniSPUJGKP17} models, where an encoder converts the source language sentence and a decoder produces the translated sentence. These methods heavily depend on large bilingual parallel corpora for training, aligning source sentences with their corresponding translations. Data augmentation\cite{sennrich2016improving,wei2023text} is a common practice in traditional machine translation. Some studies\cite{DBLP:conf/iclr/Gu0XLS18,DBLP:conf/emnlp/GhazvininejadLL19,DBLP:conf/inlg/WangGWCSWZTY21,DBLP:journals/corr/abs-2112-11640,DBLP:journals/corr/abs-2112-11642} also investigate transforming them into parallel architectures to speed up inference efficiency.

\paragraph{LLM-based Methods} In recent years, Language Model (LM)-based approaches have gained attention in the field of machine translation. These approaches leverage pre-trained language models, such as the GPT (Generative Pre-trained Transformer)\cite{DBLP:journals/corr/abs-2005-14165,DBLP:journals/jmlr/ChowdheryNDBMRBCSGSSTMRBTSPRDHPBAI23,touvron2023llama} series, and adapt them for translation tasks.

One line of LLM-based methods focuses on zero-shot or few-shot translation by incorporating in-context learning\cite{DBLP:journals/corr/abs-2302-09210}. By conditioning the LLM on a source sentence, the model can generate translations in the target language without explicitly using parallel data. This approach has shown promising results in enabling translation for language pairs with limited or no parallel resources.

Another approach involves using a small amount of high-quality bilingual parallel data to construct translation-guiding instructions. These instructions explicitly define the translation behavior by providing source-language consistent cues during the supervised fine-tuning (SFT) process. By utilizing these specially crafted instructions, the LM can be fine-tuned to perform translation more accurately and robustly.

Overall, LLM-based methods present alternative approaches to machine translation, exploring the potential of leveraging pre-trained models and incorporating limited parallel resources or high-quality instructions to enhance translation quality.

\section{A New Training Recipe}

We propose an innovative training strategy to enhance the translation capabilities of Large Language Models. As shown in Figure \ref{figure:model}, our approach consists of three stages: (1) Secondary Pre-training using Extensive Monolingual Data, (2) Continual Pre-training with Interlinear Text Format Documents, and (3) Leveraging Source-Language Consistent Instruction for Supervised Fine-Tuning.

\subsection{Stage 1: Continual Pre-training using Extensive Monolingual Data} In this stage, our aim is to enhance the training of large language models (LLMs) by utilizing diverse monolingual data. Currently, existing large models, such as Llama, are primarily pre-trained on English-centric corpora, resulting in relatively weaker comprehension and generation abilities in non-English languages. To expand the multilingual generation capabilities of LLMs, we suggest an incremental pre-training approach using extensive monolingual data. 

\textbf{It is important to note that this stage primarily focuses on enhancing the intrinsic multilingual capacity of LLMs. While it is inherently related to machine translation tasks, it is not essential.} On the one hand, we can select an existing LLM that already demonstrates robust multilingual capabilities as the base model for further training. On the other hand, even LLMs with limited multilingual support can benefit from the subsequent stages outlined in our methodology.

\subsection{Stage 2: Continual Pre-training with Sentence-aligned Parallel Data}

\begin{figure}[th]
\centering
\includegraphics[width=7.5cm]{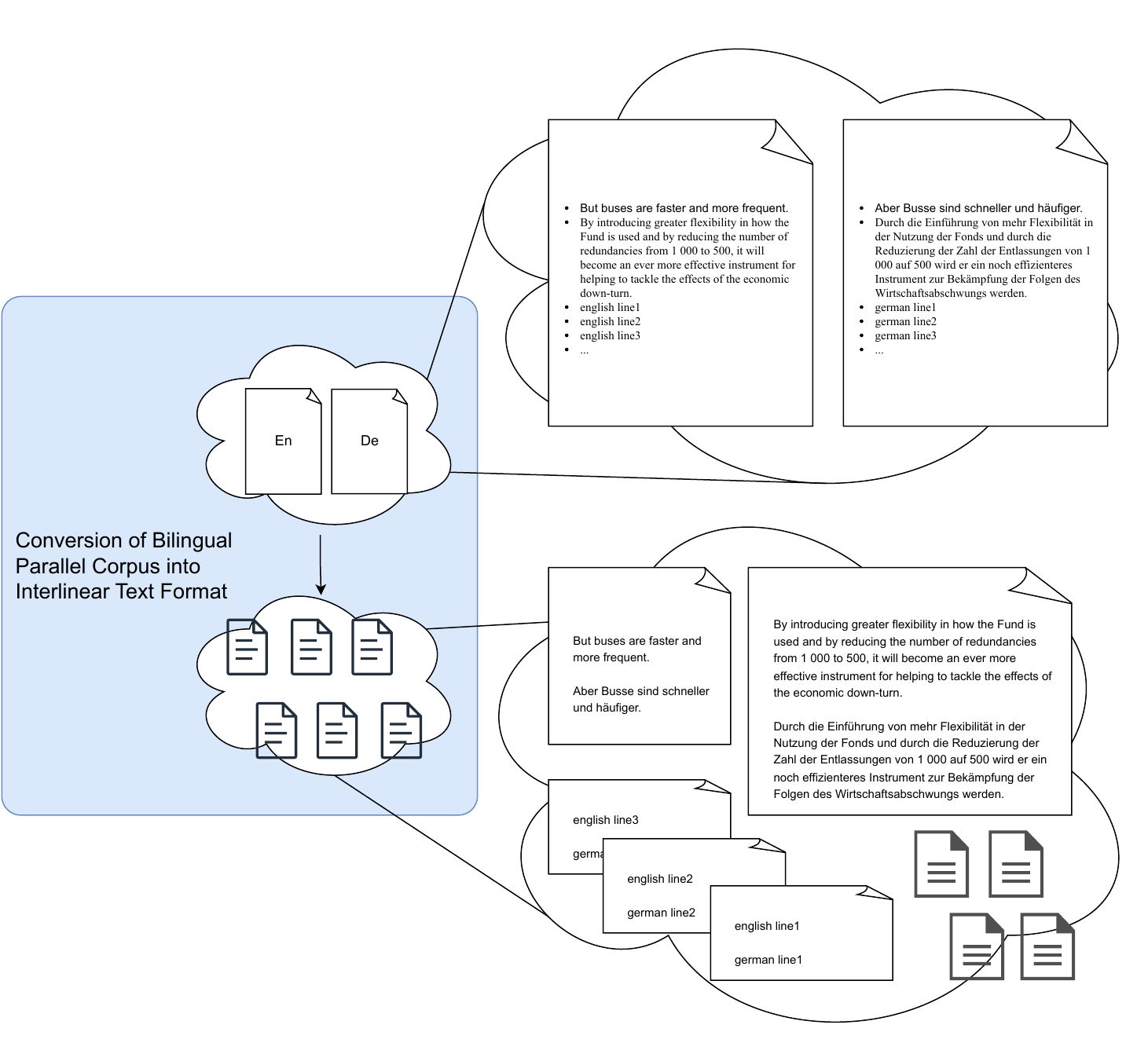}
\caption{Interlinear Text Format Documents}
\label{figure:stage2}
\end{figure}

\paragraph{Interlinear Text Format} Interlinear Text Format are a specific type of parallel text resource that consists of source sentences and their corresponding translations displayed in a aligned format. Each source sentence is accompanied by its translation, typically presented word-by-word or phrase-by-phrase, to facilitate a clear interlingual correspondence. We build the Sentence-aligned Parallel Data into this format. See Figure \ref{figure:stage2}.

Utilizing Interlinear Text Format offers several advantages for language understanding and translation tasks. Firstly, these data provide explicit linguistic alignment at a fine-grained level, enabling the model to capture syntactic and semantic correspondences across languages. This aligns closely with the goals of machine translation, as it facilitates accurate encoding of source language information and improves the quality of generated translations. Additionally, interlinear data contributes to the learning of interlingual representations, allowing the model to better understand the relationship and transferability between languages.

\paragraph{Continual Pre-training} To leverage the benefits of Interlinear Text Documents, we propose a Continual Pre-training strategy based on the LoRA\cite{hu2021lora} (Low-Rank Adaptation of Large Language Models) framework. LoRA is a robust and effective pre-training approach for language models, introduced in recent research.



By leveraging the inherent alignment information present in Interlinear Text Documents, the model learns to align and generate translations that maintain syntactic and semantic consistency with the source sentences. This continual training process allows the model to progressively improve its ability to capture cross-lingual correspondences, resulting in enhanced translation quality.

\subsection{Stage 3: Leveraging Source-Language Consistent Instruction for Supervised Fine-Tuning}



\paragraph{Source-Language Consistent Instruction} In the field of machine translation, "Source-Language Consistent Instruction" refers to the practice of constructing translation instructions that maintain consistency with the source language, aiming to achieve better results. This approach involves generating instructions that are closely related to the source language. By providing more accurate and clear guidance for supervised fine-tuning of models, this technique enhances translation quality.

\begin{figure}[th]
\centering
\includegraphics[width=6.5cm]{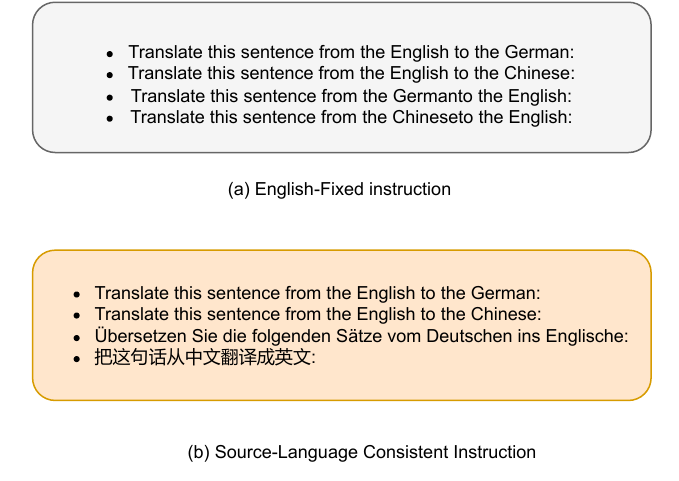}
\caption{Instruction Format}
\label{figure:stage3}
\end{figure}

To illustrate this concept, let's consider translations in English$\Leftrightarrow$Chinese and English$\Leftrightarrow$German. Traditional approaches typically employ a standardized English-Fixed instruction such as "\textit{Translate this sentence from the source language to the target language:}". However, in Source-Language Consistent Instruction, the instruction varies based on the language pair. For English-to-Chinese translation, the instruction would be "\textit{\begin{CJK*}{UTF8}{gbsn}把这句话从中文翻译成英文：\end{CJK*}}" (Translate this sentence from Chinese to English). Similarly, for German-to-English translation, the instruction would be "\foreignlanguage{german}{\textit{Übersetzen Sie die folgenden Sätze vom Deutschen ins Englische:}}" (Translate the following sentences from German to English). By utilizing language-specific instructions, there is a semantic consistency established between the instruction and the source language, resulting in clearer and more accurate guidance. As shown in Figure \ref{figure:stage3}.

\paragraph{Supervised Instruction Fine-Tuning} Supervised Instruction Fine-Tuning for machine translation tasks incorporates two pivotal aspects. Firstly, akin to the earlier phase of Continual Pre-training, we employ LoRA\cite{hu2021lora} to finely tune specific parameters of Language Learning Models (LLMs), thereby enhancing their efficiency. LoRA\cite{hu2021lora} plays a crucial role in preventing model overfitting and leads to notable performance improvements. With this approach, we judiciously fine-tune a subset of model parameters using low-rank updates, striking a delicate balance between model adaptation and computational efficiency.

Secondly, as emphasized in prior studies\cite{DBLP:journals/corr/abs-2305-11206,DBLP:conf/acl/MaillardGKSGKFG23,xu2023paradigm}, LLMs exhibit benefits from a limited yet high-quality dataset. To ensure optimal data quality during the fine-tuning process, we leverage exceptional data sources. In line with previous research, we make use of meticulously curated human-written datasets derived from the WMT test data. These datasets undergo rigorous quality control measures, rendering them an ideal choice for fine-tuning purposes.

\section{Experiments}
\subsection{Datasets and Evaluation Metrics}

The overall data statistics are shown in Table \ref{table:data-statistic} of Appendix \ref{sec:appendix_data_statistics}. For Stage 2, we utilized the WMT bilingual training dataset consisting of English$\Leftrightarrow$German (En$\Leftrightarrow$De) and English$\Leftrightarrow$Chinese (En$\Leftrightarrow$Zh) sentence pairs. The En$\Leftrightarrow$De dataset comprised approximately 4.5 million pairs, while the En$\Leftrightarrow$Zh dataset contained around 25 million pairs. Due to the higher number of En$\Leftrightarrow$Zh pairs compared to En$\Leftrightarrow$De, we sampled 4.5 million En$\Leftrightarrow$Zh pairs for our experiments. Overall, the combined dataset contained nearly 1 billion tokens.

For Stage 3, we employed the newstest2017-2020 dataset for both En$\Leftrightarrow$Zh and En$\Leftrightarrow$De translation tasks. This dataset included a total of 37.6 thousand sentence pairs for each language direction. To ensure consistency across the source language and target language, we organize these sentence pairs into Source-Language Consistent Instructions.

We additionally incorporate the test sets from the WMT22 competition, which are carefully curated to include more recent content from diverse domains such as news, social media, e-commerce, and conversations. The test sets for the De$\Rightarrow$En, En$\Rightarrow$De, Zh$\Rightarrow$En, and En$\Rightarrow$Zh tasks consist of 1984, 2037, 1875, and 2037 samples, respectively.

For automatic evaluation, we utilize SacreBLEU, which implements BLEU\cite{DBLP:conf/acl/PapineniRWZ02}, and COMET\cite{DBLP:conf/emnlp/ReiSFL20} from Unbabel/wmt22-comet-da. SacreBLEU calculates similarity based on n-gram matching, while COMET leverages cross-lingual pretrained models for evaluation.

\subsection{Setup}

We conducted our experiments using HuggingFace Transformers with open-source LLMs from the LLaMA\cite{touvron2023llama} family. Specifically, we utilized LLaMA2-7b with matched parameters as our foundation model. Additionally, we included LLaMA2-13b to explore the impact of different model sizes.

Due to computational constraints, we did not reproduce the foundational experiments from Stage 1. After Stage 1, we selected Chinese-LLaMA2\cite{Chinese-LLaMA-Alpaca} as our new foundation model. Chinese-LLaMA2 is an extended and optimized version of Llama-2, specifically tailored for Chinese language understanding and instruction comprehension. It incorporates a larger Chinese vocabulary and undergoes incremental pretraining on a large-scale Chinese dataset, which further enhances its semantic understanding capabilities.

For Stage 2, Continual Pre-training, and Stage 3, Supervised Fine-Tuning, we referred to the hyperparameters employed in the Chinese-LLaMA2 project. During Stage 2, we trained the model for 1 epoch, and for Stage 3, we extended the training to 3 epochs.

Our experiments were conducted on 8 Nvidia GPUs with 64GB of memory each, utilizing DeepSpeed\cite{DBLP:conf/kdd/RasleyRRH20} ZeRO 2 for model parallelization.

\begin{table*}[th]
\centering
\resizebox{0.98\linewidth}{!}{
\begin{tabular}{lcccccccc}
\hline
\multirow{2}*{\textbf{Models}} & \multicolumn{2}{c}{\textbf{De$\Rightarrow$En}} & \multicolumn{2}{c}{\textbf{En$\Rightarrow$De}} & \multicolumn{2}{c}{\textbf{Zh$\Rightarrow$En}} &  \multicolumn{2}{c}{\textbf{En$\Rightarrow$Zh}}\\
\cline{2-9}
 ~ & BLEU & COMET & BLEU & COMET & BLEU & COMET & BLEU & COMET \\
\hline
& \multicolumn{8}{c}{SoTA models} \\
NLLB-54B\cite{nllbteam2022language} & 26.89 & 78.94 & 34.50 & 86.45 & 16.56 & 70.70 & 27.38 & 78.91 \\
NLLB-54B Fine-tune & 27.34 & 79.86 & 35.07 & 86.95 & 17.26 & 71.35 & 27.89 & 80.13 \\
GPT-3.5-D, zero-shot & 30.90 & 84.79 & 31.80 & 85.61 & 25.00 & 81.60 & 38.30 & 85.76 \\
GPT-3.5-T, zero-shot & 33.10 & 85.50 & 34.40 & 87.00 & 26.60 & 82.90 & 44.90 & 87.00 \\
GPT-4 & 33.87 & 85.62 & 35.38 & 87.44 & 27.20 & 82.79 & 43.98 & 87.49 \\
\hline
& \multicolumn{8}{c}{Prior Similar Studies} \\
TIM-7B\cite{zeng2023tim} & 27.91 & 82.80 & 25.59 & 82.56 & 19.33 & 75.46 & 19.33 & 75.46 \\
Parrot-7B\cite{jiao2023parrot} & 29.80 & 83.00 & 26.10 & 81.60 & 20.20 & 75.90 & 30.30 & 80.30 \\
SWIE-7B\cite{chen2023improving} & 30.48 & 82.97 & 27.21 & 82.36 & 21.30 & 76.48 & 31.24 & 80.63 \\
ALMA-7B\cite{xu2023paradigm} & 29.56 & 83.95 & 30.31 & 85.59 & \textbf{23.64} & \textbf{79.78} & 36.48 & 85.05 \\
\hdashline
Parrot-13B\cite{jiao2023parrot} & 31.10 & 83.60 & 28.10 & 82.60 & 21.70 & 76.70 & 31.70 & 81.00 \\
BigTranslate-13B\cite{yang2023bigtranslate} & 23.35 & 80.68 & 21.48 & 78.81 & 14.16 & 74.26 & 28.56 & 81.31 \\
Bayling-13B\cite{zhang2023bayling} & 27.34 & 83.02 & 25.62 & 82.69 & 20.12 & 77.72 & 37.92 & 84.62 \\
ALMA-13B\cite{xu2023paradigm} & 31.14 & 84.56 & 31.47 & 85.62 & \textbf{25.46} & \textbf{80.21} & 39.84 & 85.96 \\
\hline
\textbf{Ours} & \multicolumn{8}{c}{Our Recipe with Backbone Model: LLaMA2\cite{touvron2023llama}} \\
7B Stage3 & 30.02 & 84.09 & 25.40 & 82.30 & 20.59 & 76.18 & 30.60 & 80.40 \\
7B Stage1,3* & 25.20 & 78.32 & 12.50 & 69.19 & 20.90 & 76.40 & 35.00 & 84.32 \\
7B Stage2,3 & \textbf{31.14} & \textbf{84.70} & \textbf{30.50} & \textbf{85.62} & 21.97 & 78.45 & \textbf{39.00} & \textbf{85.79} \\
7B Stage1,2,3* & 30.10 & 83.96 & 29.90 & 83.86 & \textbf{22.20} & \textbf{79.88} & \textbf{41.10} & \textbf{86.37} \\
\hdashline
13B Stage3 & 31.70 & 84.39 & 28.80 & 83.87 & 21.40 & 77.68 & 35.90 & 84.23 \\
13B Stage1,3* & 26.13 & 78.65 & 12.79 & 72.23 & 21.40 & 78.28 & 37.34 & 85.27 \\
13B Stage2,3 & \textbf{32.24} & \textbf{85.17} & \textbf{32.53} & \textbf{86.14} & 22.57 & 79.05 & \textbf{40.40} & \textbf{85.98} \\
13B Stage1,2,3* & 30.21 & 84.26 & 30.41 & 84.72 & \textbf{23.10} & \textbf{80.53} & \textbf{42.30} & \textbf{86.65} \\
\hline
\end{tabular}
}
\caption{\textbf{The overall results.} Note: * Due to computational constraints, we did not reproduce the foundational experiments from Stage 1, but instead directly utilized the Chinese-Llama2\cite{Chinese-LLaMA-Alpaca} that had undergone similar training. Since Chinese-Llama2\cite{Chinese-LLaMA-Alpaca} was only trained in Chinese during Stage 1, our main analysis about Stage 1 focuses on its performance in Zh$\Rightarrow$En and En$\Rightarrow$Zh translations.}
\label{table:main-results}
\end{table*}

\subsection{Baselines}

We evaluate our method against two baseline categories, consistent with previous studies. Firstly, we compare our approach to prior studies that share our goal of leveraging LLMs for translation. Secondly, we benchmark against the current state-of-the-art (SoTA) translation models. It's important to note that this comparison may not be entirely fair due to disparities in training data and model architectures. For example, there is a significant contrast between the 175B GPT-3.5 model and our 7B model. Nevertheless, by utilizing the same test set, we can gain insights into our model's current performance.

In the category of prior similar work, we compare our model to the following approaches: BigTranslate\cite{yang2023bigtranslate}, which extends LLaMA-1-13B to cover over 100 translation directions; TIM\cite{zeng2023tim}, which leverages correct and incorrect examples to aid LLMs in learning translation; ParroT\cite{jiao2023parrot}, through three types of instructions including translation instruction, contrastive instruction, and error-guided instruction, improves the translation performance of LLM after SFT; SWIE\cite{chen2023improving}, which enhances LLMs in translation through instruction augmentation; BayLing\cite{zhang2023bayling}, which incorporates interactive translation instructions; and ALMA\cite{xu2023paradigm}, a two-stage fine-tuning method that initially fine-tunes on monolingual data and subsequently on a small set of high-quality parallel data.

In the SoTA models category, we consider the following: the NLLB-54B\cite{nllbteam2022language} model, the largest and best translation model released in the NLLB family; the zero-shot performance of GPT3.5-text-davinci-003 (GPT-3.5-D) and GPT-3.5-turbo-0301 (GPT-3.5-T). Additionally, we present the zero-shot results for GPT-4.\footnote{GPT-3.5-D, GPT-3.5-T and GPT-4 results are sourced from \citealp{xu2023paradigm}} For a fair comparison, we also compared the results of fine-tuning NLLB-54B model with 37.6k data in Stage 3. To evaluate these baselines, we employ the same test data and evaluation metrics, reporting BLEU\cite{DBLP:conf/acl/PapineniRWZ02} and COMET\cite{DBLP:conf/emnlp/ReiSFL20} scores as provided in their respective papers.

\section{Results and Analysis}

As shown in Table \ref{table:main-results}, \textbf{overall, our results outperform all previous studies, NLLB-54B\cite{nllbteam2022language}, and GPT-3.5-D, except for a slight lag in Zh$\Rightarrow$En. Even our 7B model surpasses the results of other works. Particularly in the En$\Rightarrow$Zh direction, our BLEU score is approximately 2.5 higher than the previous state-of-the-art.} These findings are a testament to the effectiveness of our approach.

\subsection{Assessing the Impact of Stage 1}

Just as mentioned earlier, we didn't specifically train Llama2 in Stage 1, but instead directly utilized the Chinese-Llama2\cite{Chinese-LLaMA-Alpaca} that had undergone similar training. Since Chinese-Llama2\cite{Chinese-LLaMA-Alpaca} was only trained in Chinese during Stage 1, our main analysis focuses on its performance in Zh$\Rightarrow$En and En$\Rightarrow$Zh translations.

As shown in Table \ref{table:main-results}, our findings align with previous research conclusions that incremental training on monolingual data is beneficial. Furthermore, we discovered that this benefit primarily affects the target language in translation tasks. For example, we observed a significant improvement in the performance of the 7B model on the En$\Rightarrow$Zh test set, where the BLEU score increased from 30.60 to 35.00, a substantial improvement of 4.4 points. However, the improvement in the Zh$\Rightarrow$En direction was limited, indicating that the role of Stage 1 is to enhance generation rather than comprehension.

Additionally, we found that performing incremental training on only one monolingual dataset had disastrous effects on translation tasks in other languages. For example, on the En$\Rightarrow$De test set, the BLEU score plummeted from 25.40 to 12.50. Therefore, for multilingual translation, it is crucial to conduct Stage 1 training on multiple languages.

\subsection{Measuring the Effectiveness of Stage 2}

As shown in Table \ref{table:main-results}, Llama2\cite{touvron2023llama} demonstrates improved quality across various test sets after Stage 2 training. An interesting observation, considering Llama2 as a large model primarily focused on English, is that the enhancement in English-Other translations is particularly noteworthy after Stage 2 Training. For instance, the 7B model exhibits remarkable improvements in En$\Rightarrow$De, with the BLEU score increasing from 25.40 to 30.50, and in En$\Rightarrow$Zh, where it rises from 30.60 to 39.00. The magnitude of these improvements is quite significant. Encouragingly, there are also improvements observed in translations from other languages to English.

An even more intriguing finding is that, as mentioned before, since Chinese-Llama2\cite{Chinese-LLaMA-Alpaca} only underwent Stage 1 training on Chinese, its translation performance substantially deteriorates in the En$\Rightarrow$De direction. However, with the magical touch of Stage 2 training, these capabilities are miraculously restored. The 7B model, on En$\Rightarrow$De, rebounds from 12.50 to 29.90, approaching the results of the original Llama2\cite{touvron2023llama}. These outcomes effectively affirm the effectiveness of Stage 2.

After considering the overall process, we are interested in understanding the impact of Stage 2 only. As mentioned before, LLMs typically include two main types of models: Foundation Models and Chat Models. Evaluating Stage 2 essentially assesses the Foundation Model by using an n-shot evaluation, which includes both zero-shot and 5-shot evaluations. We have noticed that zero-shot evaluations can occur hallucinations. Hence, we are presenting the results of the 5-shot evaluation in Table \ref{table:stage2only}.

\begin{table}[th]
\centering
\resizebox{0.94\linewidth}{!}{
\begin{tabular}{lcccc}
\hline
Models & \multicolumn{2}{c}{Zh$\Rightarrow$En} & \multicolumn{2}{c}{En$\Rightarrow$Zh} \\
 & BLEU & COMET & BLEU & COMET \\
\hline
Baseline & 20.63 & 76.32 & 29.96 & 79.34 \\
 + Stage2 & 21.64 & 78.07 & 38.62 & 85.30 \\
\hline
\end{tabular}}
\caption{Results of the five-shot results based on Llama2-7B\cite{touvron2023llama} model.}
\label{table:stage2only}
\end{table}

\subsection{Analyzing the Outcomes of Stage 3}

To evaluate the effectiveness of Source-Language Consistent Instruction in Stage 3, we conducted a comparative experiment using English-Fixed Instruction. The results of the experiment are presented in Table \ref{table:stage3}. It is evident that in the En$\Rightarrow$De and De$\Rightarrow$En directions, the performance of these two types of instructions is quite similar. However, in the Zh$\Rightarrow$En and En$\Rightarrow$Zh directions, the use of Source-Language Consistent Instruction clearly outperforms.

\begin{table}[th]
\centering
\resizebox{0.98\linewidth}{!}{
\begin{tabular}{lcccc}
\hline
Models & De$\Rightarrow$En & En$\Rightarrow$De & Zh$\Rightarrow$En & En$\Rightarrow$Zh \\ 
\hline
Stage 3 & 30.02 & 25.40 & 20.59 & 30.60 \\
 w/o & 30.40 & 25.20 & 18.39 & 28.30 \\
\hline
Stage2,3 & 31.14 & 30.50 & 21.91 & 39.00 \\
 w/o & 31.00 & 30.23 & 18.93 & 38.69 \\
\hline
\end{tabular}}
\caption{Results of the comparative experiments based on Llama2-7B\cite{touvron2023llama} model. [w/o] means using English-Fixed Instruction.}
\label{table:stage3}
\end{table}

We believe that the similarity between English and German, as they belong to the same language family, contributes to the lack of noticeable differences. However, when dealing with cross-language pairs, employing Source-Language Consistent Instruction further enhances the translation quality.

\subsection{Comparing the Difference with ALMA}

We have noticed that our work shares some similarities with ALMA\cite{xu2023paradigm} in terms of the process, involving Continual Pre-training followed by Supervised Fine-Tuning. However, there are notable differences between our approaches.

ALMA suggests that the impact of bilingual data is reduced in the era of large models. In contrast, we firmly believe in the continued strength of bilingual data and its application in Continual Pre-training through Interlinear Text Format Documents. While ALMA acknowledges the effectiveness of conducting Continual Pre-training on monolingual data, we have also validated this finding in our own work and reached the same conclusion. However, it is important to note that our approach primarily enhances the multilingual generation capability of the large model itself, rather than being specifically tailored to translation tasks. Furthermore, ALMA utilizes a significantly larger training dataset, ranging from 13B to 20B, compared to our own.

\section{Ablation Study: What if we directly employ a large quantity of translation data for SFT?}

Both Continual Pre-training and Supervised Fine-Tuning involve incremental training on the original model. However, if we skip Stage 2 training and directly utilize the translation data from Stage 2 as instruction data for SFT, i.e., conducting SFT directly with a substantial amount of translation data, will it yield consistent improvement?

\begin{table}[th]
\centering
\resizebox{0.98\linewidth}{!}{
\begin{tabular}{ccccc}
\hline
Data Size & De$\Rightarrow$En & En$\Rightarrow$De & Zh$\Rightarrow$En & En$\Rightarrow$Zh \\ 
\hline
37.6K & 30.02 & 25.40 & 20.59 & 30.60 \\
400K & 30.20 & 25.60 & 18.49 & 31.74 \\
4,000K & 30.66 & 25.12 & 20.77 & 32.22 \\
\hline
\end{tabular}}
\caption{Results of the ablation experiments based on Llama2-7B\cite{touvron2023llama} model under different Stage 3 data size.}
\label{table:directSFT}
\end{table}

To address this question, we conducted an ablation experiment. Our Stage 3 training data consisted of 37.6k samples. Randomly selecting and merging some data from the Stage 2 training data with the Stage 3 training data, we created three sets: 37.6K, 400K, and 4,000K. The experimental results are presented in Table \ref{table:directSFT}.

We found that augmenting the training data in Stage 3 slightly improved translation quality for certain test sets. This indicates that a small amount of high-quality data is sufficient for the SFT stage.

Now, our focus is solely on the translation task. However, if we were conducting multi-task SFT, it is unlikely that other tasks would have as extensive data as machine translation. Therefore, using a large amount of translation data during SFT would result in the problem of imbalanced data distribution with other tasks. Hence, the optimal approach would still be to utilize this substantial data during the earlier stage of Continual Pre-training.

\section{Conclusions}

In this study, we have introduced a novel paradigm for enhancing the translation capabilities of large language models in machine translation tasks. Our three-stage approach, including Secondary Pre-training using Extensive Monolingual Data, Continual Pre-training with Interlinear Text Format Documents, and Leveraging Source-Language Consistent Instruction for Supervised Fine-Tuning, addresses the limitations of previous strategies and offers notable improvements in translation quality. We emphasize the significance of pre-training stages in enhancing LLMs' cross-lingual alignment abilities and the effectiveness of using a smaller but high-quality set of bilingual data during supervised fine-tuning. Notably, Stage2, which involves Continual Pre-training with Interlinear Text Format Documents, stands out as a highly efficient method, requiring minimal training data. Furthermore, aligning the instructional setting with the source language during supervised fine-tuning, as observed in Stage3, yields positive effects. The findings from this paper contribute to advancing the field of machine translation and offer valuable insights for optimizing the translation capabilities of large language models. Future research can explore additional language pairs, alternative data augmentation techniques, and different pre-training strategies to further refine our proposed paradigm.

\section{Limitations}

Despite notable contributions, this study has certain limitations. Firstly, the proposed method exhibits slightly reduced performance in the Zh$\Rightarrow$En translation direction, necessitating further analysis and improvements. Secondly, the presence of illusionary translations within large models was observed but not extensively explored. Future research should delve deeper into this phenomenon. Lastly, while the paper primarily focuses on SFT for machine translation, opportunities exist to explore SFT techniques in diverse contexts such as style translation and colloquial translation. Addressing these limitations would further enhance the effectiveness and applicability of the proposed methods.

\bibliography{acl_natbib}

\appendix

\begin{table*}[th]
\centering
\resizebox{0.92\linewidth}{!}{
\begin{tabular}{cll}
\hline
\textbf{Stage} & \multicolumn{1}{c}{\textbf{Data}} & \multicolumn{1}{c}{\textbf{Description}} \\
\hline
Stage 1 & 120G text & 120G Chinese text described in Chinese-LLaMA2\cite{Chinese-LLaMA-Alpaca} \\
\hdashline
Stage 2 & 1B tokens & 1B tokens including four directions: En$\Leftrightarrow$De and En$\Leftrightarrow$Zh.\\
 & & Each direction includes 4.5 million pairs. \\
\hdashline
Stage 3 & 37.6k pairs& 37.6k pairs combined wmt newstest2017-2020 testset with all four directions. \\
\hline
\end{tabular}
}
\caption{\textbf{The overall data statistics.} }
\label{table:data-statistic}
\end{table*}

\section{Appendix A: Data Statistics}
\label{sec:appendix_data_statistics}

Table \ref{table:data-statistic} displays the comprehensive data statistics. For Stage 2, we utilized the WMT bilingual training dataset that includes English$\Leftrightarrow$German (En$\Leftrightarrow$De) and English$\Leftrightarrow$Chinese (En$\Leftrightarrow$Zh) sentence pairs. The En$\Leftrightarrow$De dataset comprised approximately 4.5 million pairs, while for the En$\Leftrightarrow$Zh dataset, we randomly sampled an equivalent number of pairs from the total 25 million pairs. In total, the combined dataset contained close to 1B tokens.

Moving on to Stage 3, we utilized the newstest2017-2020 dataset for both the En$\Leftrightarrow$De and En$\Leftrightarrow$Zh translation tasks. This dataset comprised 37.6 thousand sentence pairs for each language direction. To maintain coherence between the source and target languages, we categorized these sentence pairs into Source-Language Consistent Instructions.

\section{Appendix B: Stage 2 Training Data Sample}
\label{sec:appendix_stage_2}

\begin{figure*}[th]
\centering
\includegraphics[width=9cm]{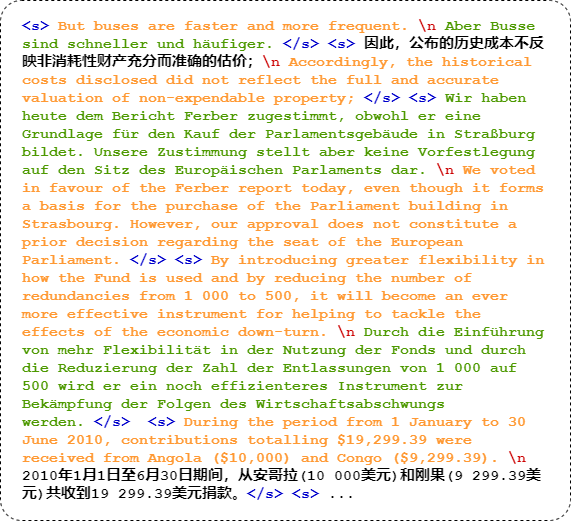}
\caption{Stage 2 Training Data Sample}
\label{figure:stage2_data}
\end{figure*}

Figure \ref{figure:stage2_data} displays samples of the Stage 2 training data.

\end{document}